\documentclass[10pt,twocolumn,letterpaper]{article}

\usepackage[pagenumbers]{cvpr} 

\usepackage{graphicx}
\usepackage{amsmath}
\usepackage{amssymb}
\usepackage{booktabs}
\usepackage{tabularx}
\usepackage{makecell}
\usepackage{chngcntr}
\usepackage[accsupp]{axessibility}
\usepackage[super]{nth}

\newcolumntype{Y}{>{\centering\arraybackslash}X}
\newcommand{\norm}[1]{\left\lVert#1\right\rVert}

\usepackage[pagebackref,breaklinks,colorlinks]{hyperref}

\usepackage[capitalize]{cleveref}
\crefname{section}{Sec.}{Secs.}
\Crefname{section}{Section}{Sections}
\Crefname{table}{Table}{Tables}
\crefname{table}{Tab.}{Tabs.}

\def\cvprPaperID{6419} 
\def\confName{CVPR}
\def\confYear{2023}

\begin{document}

\title{Unsupervised Contour Tracking of Live Cells \\by Mechanical and Cycle Consistency Losses}

\author{\qquad Junbong Jang \qquad \qquad \\
\qquad KAIST \qquad \qquad \\
\and
Kwonmoo Lee* \\
Boston Children's Hospital \\
Harvard Medical School
\and
Tae-Kyun Kim* \\
KAIST \\
Imperial College London 
}
\maketitle

\begin{abstract}
   Analyzing the dynamic changes of cellular morphology is important for understanding the various functions and characteristics of live cells, including stem cells and metastatic cancer cells.
   To this end, we need to track all points on the highly deformable cellular contour in every frame of live cell video. 
   Local shapes and textures on the contour are not evident, and their motions are complex, often with expansion and contraction of local contour features.
   The prior arts for optical flow or deep point set tracking are unsuited due to the fluidity of cells, and previous deep contour tracking does not consider point correspondence.
   We propose the first deep learning-based tracking of cellular (or more generally viscoelastic materials) contours with point correspondence by fusing dense representation between two contours with cross attention.
   Since it is impractical to manually label dense tracking points on the contour, unsupervised learning comprised of the mechanical and cyclical consistency losses is proposed to train our contour tracker.
   The mechanical loss forcing the points to move perpendicular to the contour effectively helps out.  
   For quantitative evaluation, we labeled sparse tracking points along the contour of live cells from two live cell datasets taken with phase contrast and confocal fluorescence microscopes. Our contour tracker quantitatively outperforms compared methods and produces qualitatively more favorable results. Our code and data are publicly available at {\small \url{https://github.com/JunbongJang/contour-tracking/}}
\end{abstract}
{\let\thefootnote\relax\footnote{* Corresponding authors: kimtaekyun@kaist.ac.kr and \\ kwonmoo.lee@childrens.harvard.edu
}}\setcounter{footnote}{0}

\begin{figure}[t]
  \centering
   \includegraphics[width=1.0\linewidth]{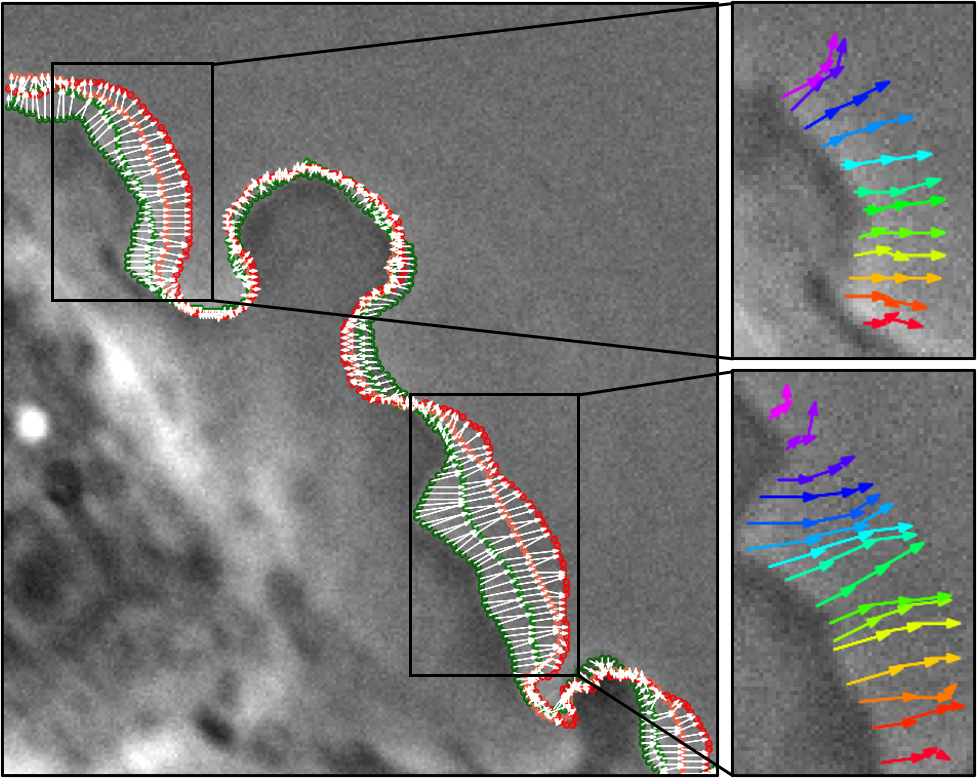}
   \caption{\textbf{Visualization of contour tracking results.} Dense point correspondences between adjacent contours are shown with white arrows overlaid on the first frame. The first frame's contour points are in dark green, and the last frame's contour points are in red. Only half of the contour points and correspondences are shown for visualization purposes. The trajectories of a few tracked points are shown on the right.}
   \label{fig:contour_tracking-fig}
   \vspace{-0.4cm}
\end{figure}
\section{Introduction}
\label{sec:intro}
During cell migration, cells change their morphology by expanding or contracting their plasma membranes continuously like viscoelastic materials \cite{Machacek2006}.
The dynamic change in the morphology of a live cell is called cellular morphodynamics and ranges from cellular to the subcellular movement of contour at varying spatiotemporal scales. While cellular morphodynamics plays a vital role in angiogenesis, immune response, stem cell differentiation, and cancer invasiveness \cite{choi2021emerging, LEE201537}, it is challenging to understand the various functions of cellular morphodynamics because its uncharacterized heterogeneity could mask crucial mechanistic details.
As an initial step to understanding cellular morphodynamics, cellular morphodynamics is quantified by tracking every point along the cellular contour (contour tracking) and estimating their velocity \cite{Machacek2006, MARS_Net, jang2022protocol}.
Then, quantification of cellular morphodynamics is further processed by other downstream machine learning tasks to characterize the drug-sensitive morphodynamic phenotypes with distinct molecular mechanisms \cite{Wang2018, ma2018profiling, choi2021emerging}. 
Because contour tracking (e.g., \cref{fig:contour_tracking-fig}) is the important first step, the tracking accuracy is crucial in this live cell analysis.

There are two main difficulties involved with contour tracking of a live cell. First, the live cell's contour exhibits visual features that can be difficult to distinguish by human eyes, meaning that a pixel and its neighboring pixels have similar color values or features. 
Optical flow \cite{Jonschkowski2020matters, sun2018pwc} can track every pixel in the current frame by assuming that the corresponding pixel in the next frame will have the same distinct feature, but this assumption is not sufficient to find corresponding pixels given cellular visual features.
Second, the expansion and contraction of the cellular contour change the total number of tracking points due to one point splitting into many points or many points converging into one. 
PoST\cite{nam2021polygonal} tracks a fixed number of a sparse set of points that cannot accurately represent the fluctuating shape of the cellular contour. 
Other deep contour tracking or video segmentation methods \cite{ yeap2017automatic, elmahdy2019robust, saboo2020deep} do not provide dense point-to-point correspondence information between a contour and its next contour.

Previous cellular contour tracking method (mechanical model) \cite{Machacek2006} evades the first problem by taking the segmentation of the cell body as inputs instead of raw images. Then, it finds the dense correspondences of all points between two contours by minimizing the normal torsion force and linear spring force with the Marquard-Levenberg algorithm \cite{more1978levenberg}.
However, the mechanical model has limited accuracy because it does not consider visual features in raw images. 
Also, its linear spring force which keeps every distance between points the same is less effective during the expansion and contraction of the cell, as shown in our experiments (see \cref{tab:loss ablation table}). 

\begin{figure*}[t]
  \centering

  \includegraphics[width=1.0\linewidth]{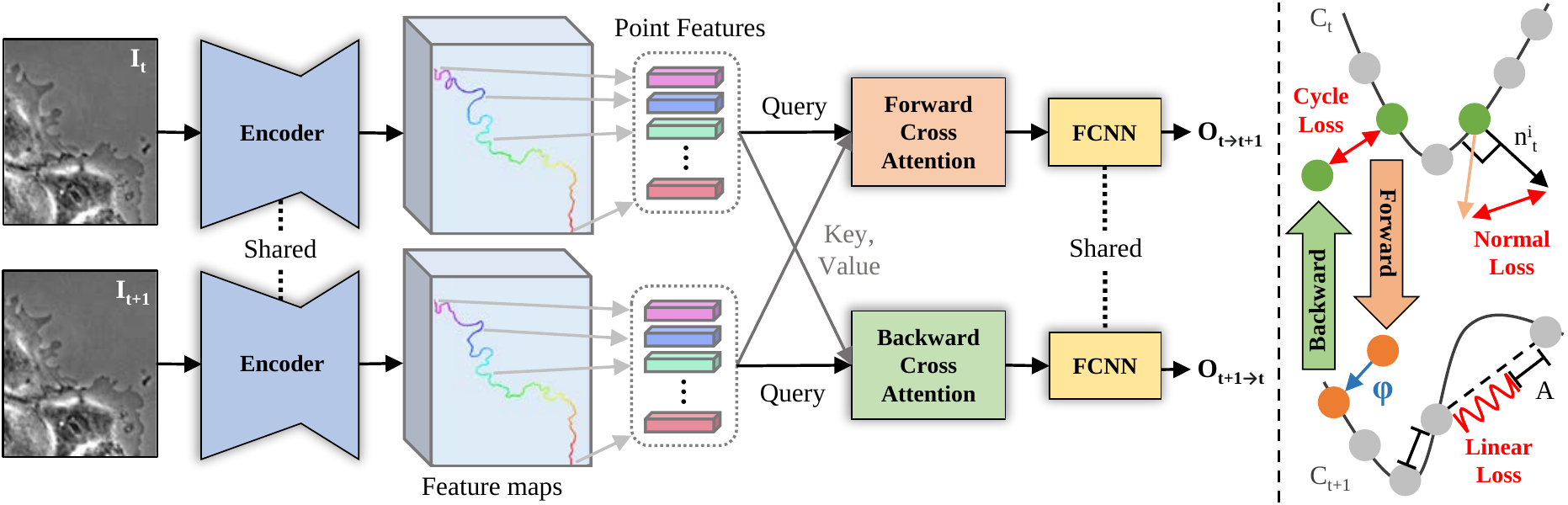}
  \caption{\textbf{Our architecture on the left and unsupervised learning losses on the right.} Shared encoder comprised of VGG16 and FPN encodes first and second images. Point features are sampled at the location of ordered contour points indicated by rainbow colors from red to purple. Point features are inputted as query or key and value to the cross attentions. Lastly, shared FCNN takes the fused features and regresses forward $O_{t \rightarrow t+1}$ or backward $O_{t+1 \rightarrow t}$ offsets. The cycle consistency, mechanical-normal, and mechanical-linear losses are shown in red color. }
  \label{fig:architecture-fig}
\end{figure*}

Therefore, we present a deep learning-based contour tracker that can overcome these difficulties.
Our contour tracker is comprised of a feature encoder, two cross attentions \cite{vaswani2017attention}, and a fully connected neural network (FCNN) for offset regression, as shown in \cref{fig:architecture-fig}.
Given two consecutive images and their contours represented as a sequence of points, our contour tracker encodes the visual features of two images and samples their feature at the location of contours.
The sampling makes our contour tracker focus on contour features and reduces the noise from irrelevant features unlike optical flow \cite{Jonschkowski2020matters}.
The cross attention \cite{vaswani2017attention} fuses the sampled features from two contours globally and locally and regresses the offset for each contour point of the first frame. 
To obtain the dense point-to-point correspondences between the current and the next contours, offset points from the current contour are matched with the closest contour points in the next frame. 
In every frame, some contour points merge due to contraction, so new contour points emerge in the next frame as shown in \cref{fig:contour_tracking-fig}. With dense point-to-point correspondences, new contour points in the next contour are also tracked.
The proposed architectural design achieves the best accuracy among variants, including circular convolutions \cite{peng2020deep}, and correspondence matrix \cite{casey2021animation}. 
To the best of our knowledge, this is the first deep learning-based contour tracking with dense point-to-point correspondences for live cells.

In this contour tracking, supervised learning is not feasible because it is difficult to label every point of the contour manually. 
Instead, we propose to train our contour tracker solely by unsupervised learning comprised of mechanical and cycle consistency losses. 
Inspired by the mechanical model \cite{Machacek2006} that minimizes the normal torsion and linear spring force, we introduce the mechanical losses to end-to-end learning. The mechanical-normal loss that keeps the angle difference small between the offset point and the direction normal to the cellular contour played a significant role in boosting accuracy. Also, we implement cycle consistency loss to encourage all contour points tracked forward-then-backward to return to their original location.
However, previous approaches such as PoST \cite{nam2021polygonal} and Animation Transformer (AnT) \cite{casey2021animation} rely on supervised learning in addition to cycle consistency loss or find mid-level correspondences \cite{wang2019learning} instead of pixel-level correspondences.

We evaluate our contour tracker on the live cell dataset taken with a phase contrast microscope \cite{MARS_Net} and another live cell dataset taken with a confocal fluorescence microscope \cite{Wang2018}.
For a quantitative comparison of contour tracking methods, we labeled sparse tracking points on the contour of live cells for all sampled frames. In total, we labeled 13 live cell videos for evaluation. Evaluation with a sparse set of points is motivated by the fact that if tracking of dense contour points is accurate, tracking any one of contour points should be accurate also. 
We also qualitatively show our contour tracker works on another viscoelastic organism, jellyfish \cite{skorokhodov2022stylegan}. 
Our contributions are summarized as follows.
\begin{itemize}
\item We propose the first deep learning-based model that tracks cellular contours densely while surpassing the accuracy of other methods.
\item We present an unsupervised learning strategy by mechanical loss and cycle consistency loss for contour tracking.
\item We demonstrate that the use of forward and backward cross attention with cycle consistency has a synergistic effect on finding accurate dense correspondences.
\item We label tracking points in the live cell videos and quantitatively evaluate cellular contour tracking for the first time.
\end{itemize}

\section{Related Work}
\label{sec:Related Work}

\subsection{Tracking}
Cell tracking is a method that tracks the movement of a cell as one object \cite{ulman2017objective}. It first segments cells and then finds their trajectories and velocities by solving the graph of potential cell tracks with integer linear programming\cite{scherr2020cell}. 
Another work uses coupled minimum-cost flow \cite{padfield2011coupled} to account for splitting and merging events of the cell. 
The entire population of cells can be densely tracked by the optical flow \cite{zhou2019motion}.
In contrast, our contour tracker deals with points along the cellular contour, which is a portion of the entire cell.

The optical flow can be used for contour tracking since it can track the movement of every pixel in the video \cite{balasundaram2019optical}.
When there is no ground truth optical flow, an unsupervised optical flow model such as UFlow\cite{Jonschkowski2020matters} is used.
UFlow is based on the PWC-Net \cite{sun2018pwc} and trains by minimizing the photometric loss between the warped second image and the first image in pixel intensity.
However, optical flow can be confused by tracking features that are not related to contour. Occasionally, visible membrane features go inside cellular interiors, hindering accurate contour tracking.

Another way to perform contour tracking is by iteratively running the contour prediction method in every frame of the video. 
The contour prediction method represents the object boundary by a sequence of sparse points connected with straight lines and regresses offsets for the initial set of points to fit the object boundary.
There are conventional contour prediction methods such as Snake \cite{kass1988snakes} or deep learning-based models such as DeepSnake \cite{peng2020deep} and DANCE \cite{liu2021dance}, which performs real-time objection detection.
PoST \cite{nam2021polygonal} extends the contour prediction method to regress offsets for 128 points along the contour of an object in the current frame to get a new contour in the next frame.
It is trained by supervised learning on the synthetic dataset having distinct visual features that are easily trackable by human eyes.
Unlike PoST\cite{nam2021polygonal}, our contour tracker tracks a varying number of points along the entire contour with challenging visual features.

\subsection{Mechanical Model}
The mechanical model \cite{Machacek2006} optimizes the nonlinear equation comprised of the normal torsion force which encourages the points to move perpendicular to the contour and the linear spring force which keeps the distance between neighboring points the same.
Features in the direction normal to the contour are widely used by the active shape model \cite{cootes2000introduction}, HMM contour tracker\cite{chen2006multicue} or particle filter-based contour tracker \cite{cao2019visual} to find the matching point in the next contour. 
The linear spring force is similar to the first-order term in the Snake algorithm \cite{kass1988snakes} which minimizes the distance between points. 
Also, ant colony optimization \cite{van2007contour} improved the contour correspondence accuracy by incorporating the proximity information of neighboring points. 

\subsection{Dense Correspondences}
Dense point correspondences between the current contour and the next contour are necessary to track each contour point throughout the video.
Deformable surface tracking \cite{wang2019deformable, hilsmann2008tracking} finds the correspondences between a fixed number of key points on a fixed 3D surface area throughout the video by a mesh-based deformation model.
Animation Transformer \cite{casey2021animation} uses cross attention \cite{vaswani2017attention} to find dense correspondences between two line segments. The dense correspondences are predicted as a 2D correspondence matrix, similar to the feature matching step in the 3D point cloud registration \cite{van2011survey, min2021distinctiveness}.
ContourFlow \cite{di2015contour} finds the point correspondences among the fragmented contours.
Instead of predicting correspondences between two contours by feature matching, our contour tracker predicts the offset from current contour points to utilize mechanical loss \cite{Machacek2006}. If our contour tracker predicts the correspondence instead of offset, the computation of mechanical loss becomes non-differentiable for end-to-end learning.

\section{Method}
In this section, we explain our architecture and unsupervised learning strategy with mechanical and cycle consistency losses as shown in \cref{fig:architecture-fig}.

\subsection{Architecture}
As input, our contour tracker takes the current frame $I_t$ and contour $C_{t}$ and the next frame $I_{t+1}$ and contour $C_{t+1}$. 
The contour is comprised of a sequence of contour points $p_t^i \in C_{t}$  in 2D coordinates. 
Every pixel along the contour extracted from the segmentation mask becomes a contour point $p_t^i$.
The current frame $I_t$ and the next frame $I_{t+1}$ are encoded by ImageNet \cite{deng2009imagenet} pre-trained VGG16 \cite{SimonyanZ14a} and upsampled by Feature Pyramid Network (FPN) \cite{lin2017feature} to match the size of input images, $I_t$ and $I_{t+1}$.
Then, image features at the location of contour points are sampled from FPN feature map. The first image's features are sampled at the location of first contour points $C_{t}$ and the second image's features are sampled at the location of second contour points $C_{t+1}$.
The sinusoidal positional embedding \cite{vaswani2017attention} and contour points' coordinates are concatenated to image features.

The multi-head cross attention (MHA) \cite{vaswani2017attention} is used to fuse the feature representation of two contours with arbitrary contour lengths and to capture global and local contour features.  
Our contour tracker has forward and backward cross attentions. 
The forward cross attention takes the first contour's feature as a query and the second contour's feature as key and value.
The backward cross attention takes the second contour's feature as a query and the first contour's feature as a key and value.
Lastly, FCNN comprised of 3 linear layers with ReLU activation in between receives the fused features from the forward cross attention and regresses the offset $O_{t\rightarrow{t+1}}$ of all contour points $C_{t}$ in the first frame. 
The same FCNN receives the fused features from the backward cross attention and regresses the backward offset $O_{t+1\rightarrow{t}}$ of all contour points $C_{t+1}$ in the second frame. 
The backward offset is necessary to compute cycle consistency loss.

\subsection{Unsupervised Learning}

\noindent \textbf{Cycle Consistency Loss.}\quad
Computing cycle consistency loss is a three-step process given two consecutive images with contour points. 
First, contour points in the first frame $C_t$ move by regressing offset $O_{t\rightarrow{t+1}}$ from their current positions. 
Second, each offset point is matched with the closest point on the second frame's contour $C_{t+1}$. This operation is denoted by $\phi$.
Lastly, offset points matched to the second frame's contour are tracked back to the first frame by regressing backward offsets $O_{{t+1}\rightarrow{t}}$.
Without the second step, learning by cycle consistency loss fails because the model can regress zero forward and backward offsets to obtain zero cycle consistency loss.
$\phi$ is non-differentiable, but gradient flows to both forward and backward cross attentions because cycle consistency loss is comprised of forward and backward consistency defined as follows:
\begin{equation}
  L_{forward}=\sum_{i=0}^{N_t-1} \| p^i_t - O_{t+1\rightarrow t}(\phi(O_{t\rightarrow t+1}(p^i_t))) \|_2
  \label{eq:cyclical loss1}
\end{equation}
\begin{equation}
L_{backward}=\sum_{i=0}^{N_t-1} \| p^i_{t+1} - O_{t\rightarrow t+1}(\phi(O_{t+1\rightarrow t}(p^i_{t+1}))) \|_2
  \label{eq:cyclical loss2}
\end{equation}
\begin{equation}
L_{cycle}= L_{forward} + L_{backward}
  \label{eq:cyclical loss3}
\end{equation}
where $N_t$ denotes the total number of contour points at time $t$ because
the contour can expand or contract.

\noindent \textbf{Mechanical-Normal Loss.}\quad
We reformulate the normal force in the mechanical model \cite{Machacek2006} as follows. For each contour point, normal vectors $n^i_t$ orthogonal to the contour $C_t$ are numerically computed by approximating tangent vectors at each contour point by central difference and rotating tangent vectors by 90 degrees. Then, normal vectors and offsets of all contour points are normalized to unit vectors.
Lastly, we compute the L1 difference between them as follows:
\begin{equation}
  L_{\text{mech-normal}} = \sum_{i=1}^{N_t-2} \norm{ \frac{n^i_t}{\| n^i_t \|_2 } - \frac{ O_{t \rightarrow t+1} (p^i_t) }{ \| O_{t \rightarrow t+1} (p^i_t) \|_2 } }_1
  \label{eq:mechanical loss}
\end{equation}
From $N_t$ contour points, the first and last point is excluded from computation since their tangent vectors cannot be approximated. 

Please refer to the supplementary section for other unsupervised learning losses.
For optimal performance (see the ablation study in \cref{tab:loss ablation table}), the total training loss is a sum of the cycle consistency and mechanical-normal loss:
\begin{equation}
  L_{total} = L_{cycle} + L_{\text{mech-normal}}
  \label{eq:total loss}
\end{equation}

\subsection{Differentiable Sampling}
\begin{figure}[t]
  \centering
  \includegraphics[width=1.0\linewidth]{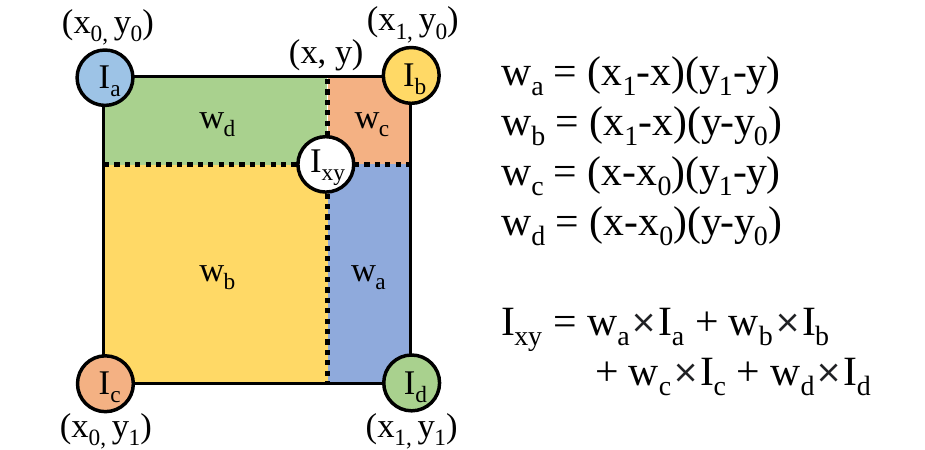}
    \vspace{-0.6cm}
   \caption{\textbf{Bilinear Sampling} of a pixel or feature at a coordinate (x,y) involves the bilinear interpolation of pixels or features at four adjacent points ($I_a$, $I_b$, $I_c$, $I_d$) in a grid by the weights ($w_a$, $w_b$, $w_c$, $w_d$).}
    \vspace{-0.4cm}
   \label{fig:sampling}
\end{figure}

To update our network during backpropagation through the sampling, we use the bilinear sampling from UFlow \cite{Jonschkowski2020matters} to retrieve the pixel intensity or image feature $I_{xy}$ at a coordinate ($x$, $y$) as shown in \cref{fig:sampling}. $I_{xy}$ has nonzero gradients with respect to coordinates or features at four adjacent points.
This method samples features located at the contour points C$_t$/C$_{t+1}$ or offset points in the training and inference.

\subsection{Pre-processing and Labeling}
\begin{figure}[b]
  \centering
   \vspace{-0.3cm}
  \includegraphics[width=1.0\linewidth]{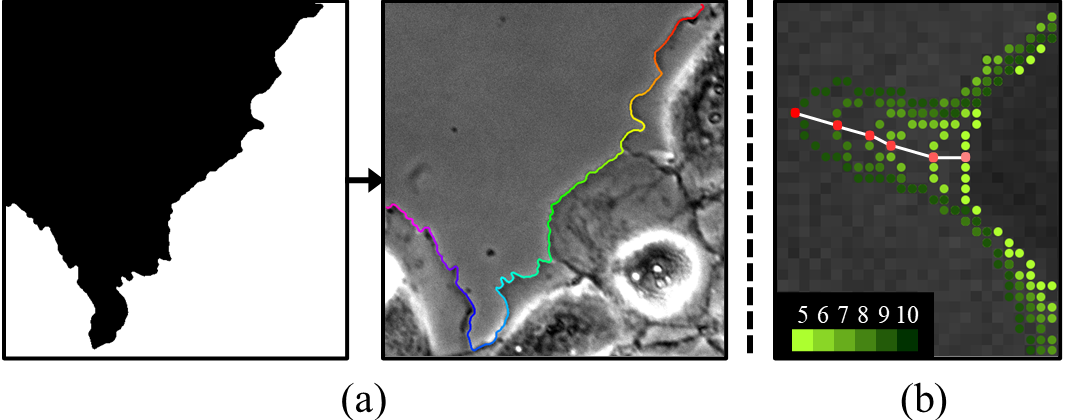}
   \vspace{-0.6cm}
   \caption{ \textbf{(a) Extraction of contour points} from the segmentation mask of one of the phase contrast live cell images \cite{MARS_Net}. Contour points are in sequential order as shown in color, from pink to red.
    \textbf{(b) Labeling tracking points} in 5x higher temporal resolution. The red point is tracked with correspondences shown in white lines. The color of the contour points changes from yellow-green to dark green as the frame number increases from 5 to 10.
    }
   \label{fig:preprocessing_labeling}
   \vspace{-0.1cm}
\end{figure}

From the binary segmentation mask, the contour with one-pixel width is extracted and the contour is converted to an ordered sequence of contour points by an off-the-shelf algorithm of contour finder \cite{opencv_library}, as shown in \cref{fig:preprocessing_labeling}(a). 
For instance, if the live cell is anchored to the left image border, the points are ordered starting from the leftmost top point in every frame. 
Points touching the image border are not considered for contour tracking.
These ordered sequences of contour points do not have point-to-point correspondences. 

For quantitative evaluation, we labeled five tracking points roughly equal distances apart from each other in low temporal resolution (every fifth frame of the video). 
However, we examined the video in 5x higher temporal resolution (every consecutive frame). 
For instance, to label the tracking point in the \nth{10} frame from the \nth{5} frame, we examined consecutive frames between the \nth{5} and the \nth{10} frames, as shown in \cref{fig:preprocessing_labeling}(b).
Labeling the tracking point can be ambiguous without examing those consecutive frames, given the large cellular motion. 
The tracking points that move outside the image boundary or become occluded due to cellular movement were not labeled.
Labeled tracking points were used for evaluation only.

\begin{table*}
  \centering
{
\setlength{\tabcolsep}{0.2em}
\renewcommand{\arraystretch}{1.0}
  \begin{tabularx}{\textwidth}{YYYYY|YYY|YYY}
    \toprule
    Supervised & Cycle & Photo & Normal & Linear & SA$_{.02}$ & SA$_{.04}$ & SA$_{.06}$ & CA$_{.01}$ & CA$_{.02}$ & CA$_{.03}$ \\
    \midrule
    
    \checkmark & & & & & 0.549 & 0.813 & 0.904 & 0.614 & 0.821 & 0.900 \\
    & \checkmark & & & & 0.632 & \underline{0.869} & \underline{0.953} & \underline{0.676} & 0.849 & \underline{0.925} \\
    & & \checkmark & & & 0.198 & 0.383 & 0.471 & 0.234 & 0.402 & 0.489 \\
    & & & \checkmark & & \underline{0.640} & 0.858 & 0.948 & 0.674 & \underline{0.853} & 0.922 \\
    & & & \checkmark & \checkmark & 0.378 & 0.593 & 0.686 & 0.400 & 0.594 & 0.674 \\
    & \checkmark & & \checkmark & \checkmark & 0.426 & 0.611 & 0.738 & 0.469 & 0.627 & 0.710\\
    & \checkmark & & \checkmark & & \textbf{0.729} & \textbf{0.937} & \textbf{0.974} & \textbf{0.762} & \textbf{0.925} & \textbf{0.971} \\
    \bottomrule
  \end{tabularx}
  
  \caption{Ablation Studies of Loss functions on phase contrast live cell videos \cite {MARS_Net}. The Cycle refers to cycle consistency loss, Photo refers to photometric loss, Normal refers to mechanical-normal loss, and Linear refers to mechanical-linear loss.}
  \label{tab:loss ablation table}
}
\end{table*}

\begin{table}
  \centering
{ \small
\setlength{\tabcolsep}{0.2em}
\renewcommand{\arraystretch}{1.0}
  \begin{tabularx}{\columnwidth}{>{\centering}m{2.3cm}|YYY|YYY}
    \toprule
    Method & SA$_{.02}$ & SA$_{.04}$ & SA$_{.06}$ & CA$_{.01}$ & CA$_{.02}$ & CA$_{.03}$ \\
    \midrule
    No Cross & 0.659 & 0.858 & 0.969 & 0.696 & 0.851 & 0.939 \\
    Single Cross & 0.677 & 0.864 & 0.930 & 0.734 & 0.854 & 0.913 \\
    Circ Conv & 0.643 & \underline{0.931} & \textbf{0.983} & 0.718 & \underline{0.910} & \underline{0.965} \\
    1D Conv & \underline{0.692} & 0.909 & \underline{0.976} & \underline{0.736} & 0.881 & 0.948 \\
    \midrule
    Ours & \textbf{0.729} & \textbf{0.937} & 0.974 & \textbf{0.762} & \textbf{0.925} & \textbf{0.971} \\
    \bottomrule
  \end{tabularx}
  
  \caption{Ablation studies of Architecture on phase contrast live cell videos \cite {MARS_Net}. }
  \label{tab:architecture ablation table}
}
\end{table}

\begin{table}
  \centering
{ \small
\setlength{\tabcolsep}{0.2em}
\renewcommand{\arraystretch}{1.0}
  \begin{tabularx}{\columnwidth}{>{\centering}m{2.3cm}|YYY|YYY}
    \toprule
    Method & SA$_{.02}$ & SA$_{.04}$ & SA$_{.06}$ & CA$_{.01}$ & CA$_{.02}$ & CA$_{.03}$ \\
    \midrule
    UFlow\cite{Jonschkowski2020matters} & 0.585 & 0.809 & 0.881 & 0.632 & 0.802 & 0.857 \\
    PoST\cite{nam2021polygonal} & 0.629 & 0.850 & \underline{0.947} & 0.693 & \underline{0.872} & \underline{0.939} \\
    Mechanical\cite{Machacek2006} & \underline{0.683} & \underline{0.853} & 0.938 & \underline{0.722} & 0.863 & 0.927 \\
    \midrule
    Ours & \textbf{0.729} & \textbf{0.937} & \textbf{0.974} & \textbf{0.762} & \textbf{0.925} & \textbf{0.971} \\
    \bottomrule
  \end{tabularx}
  
  \caption{Comparison with other methods on phase contrast live cell videos \cite {MARS_Net}.}
  \label{tab:MARS-Net table}
}
\end{table}

\begin{table}
  \centering
{ \small
\setlength{\tabcolsep}{0.2em}
\renewcommand{\arraystretch}{1.0}
  \begin{tabularx}{\columnwidth}{>{\centering}m{2.3cm}|YYY|YYY}
    \toprule
    Method & SA$_{.02}$ & SA$_{.04}$ & SA$_{.06}$ & CA$_{.01}$ & CA$_{.02}$ & CA$_{.03}$ \\
    \midrule
    UFlow\cite{Jonschkowski2020matters} & 0.605 & 0.785 & 0.863 & 0.520 & 0.685 & 0.791 \\
    PoST\cite{nam2021polygonal} & \underline{0.614} & 0.805 & \textbf{0.888} & \underline{0.531} & 0.706 & \underline{0.807} \\
    Mechanical\cite{Machacek2006} & 0.603 & \underline{0.798} & 0.876 & 0.517 & \underline{0.711} & 0.804 \\
    \midrule
    Ours & \textbf{0.632} & \textbf{0.824} & \underline{0.882} & \textbf{0.555} & \textbf{0.728} & \textbf{0.826} \\
    \bottomrule
  \end{tabularx}
  
  \caption{Comparison with other methods on confocal fluorescence live cell videos \cite{Wang2018}.}
  \label{tab:HACKS table}
}
\end{table}


\section{Experiments}
\subsection{Dataset}
The confocal fluorescence dataset \cite{Wang2018} contains 40 live cell videos taken with confocal fluorescence microscopy. 
They are classified into 9 different categories based on their cellular morphodynamics. Therefore, we randomly picked one video from each category to validate our contour tracker on all types of cellular morphodynamics. We trained on 31 live cell videos and validated our contour tracker on the other 9 live cell videos.
Phase Contrast dataset \cite {MARS_Net} contains 4 multi-cellular live cell videos and 5 single-cellular live cell videos taken with a phase contrast microscope. 
We train on 5 single-cellular live cell videos and validate our contour tracker on 4 multi-cellular live cell videos.

Each live cell video is 200 frames long, and every frame is segmented in both datasets. We sampled every fifth frame from the video for contour tracking. By sampling, we use fewer frames for contour tracking and evaluate the robustness of our contour tracker in a low temporal resolution setting. 
For labeling tracking points, we examined all 200 frames to see each tracked point's cellular topology, visual features, and trajectory from previous consecutive frames.

\subsection{Implementation Details}
For training, live cell videos from the phase contrast dataset \cite{MARS_Net} are resized to $256^2$, and the live cell videos from the confocal fluorescence dataset \cite{Wang2018} are resized to $512^2$.
We trained our contour tracker using Adam \cite{KingmaB14} optimizer with an initial learning rate of 0.0001 and linear learning rate decay after 10k iterations. 
Also, we trained our contour tracker for 50k iterations with a batch size of 8 on one TITAN RTX GPU for 1 day.
For inference, only the forward cross attention is used to regress offsets. The offset points are moved to the closest contour point by operation $\phi$ at each frame to obtain the dense correspondences between the current contour $C_t$ and the next contour $C_{t+1}$.  

\subsection{Evaluation Metrics}
To evaluate the point tracking, we use the spatial accuracy (SA) introduced in \cite{nam2021polygonal} and introduce contour accuracy (CA). SA measures the distance between the ground truth points $g_t$ and the predicted points $p_t=\phi(O_{t-1\rightarrow t}(p_{t-1}))$. If the distance is less than a threshold $\tau$, 1 is added. Otherwise, 0 is added. Each tracking point's x and y coordinates are normalized by image height and width such that the x and y coordinates range from 0 to 1. The contour points in the first frame $p_0=C_{0}$ are tracked and evaluated against the ground truth points $g_t$ at each time step t:
\begin{equation}
  SA_\tau (p_t,g_t) = \lambda \sum_{t=1}^{T-1} \sum_{i=0}^{N-1} (\| p_t^i - g_t^i \|_2 < \tau)
  \label{eq:SA}
\end{equation}
where $T$ is the total number of frames in the video, $N=5$ is the total number of labeled tracking points and $\lambda = \frac{1}{(T-1)N}$.
CA measures the arc length between two points on the contour. It is equivalent to measuring the difference between the ground truth point's indices and the predicted point's indices for contour points. Due to the fluctuating shape of the cellular contour, two points close to each other in the image space can be far apart in terms of the arc length. The arc length and spatial distance between two points are equal when the contour is a straight line. 
Let $C_t()$ be a function that returns the index of the contour point given the coordinate of the contour point.

\begin{equation}
  CA_\tau (p_t,g_t) = \lambda \sum_{t=1}^{T-1} \sum_{i=0}^{N-1} (\| C_t(p_t^i) - C_t(g_t^i) \|_1 < \tau)
  \label{eq:CA}
\end{equation}

Then, CA is normalized by the total number of contour points in the current frame. 


\begin{figure*}[t]
  \centering
   \includegraphics[width=1.0\linewidth]{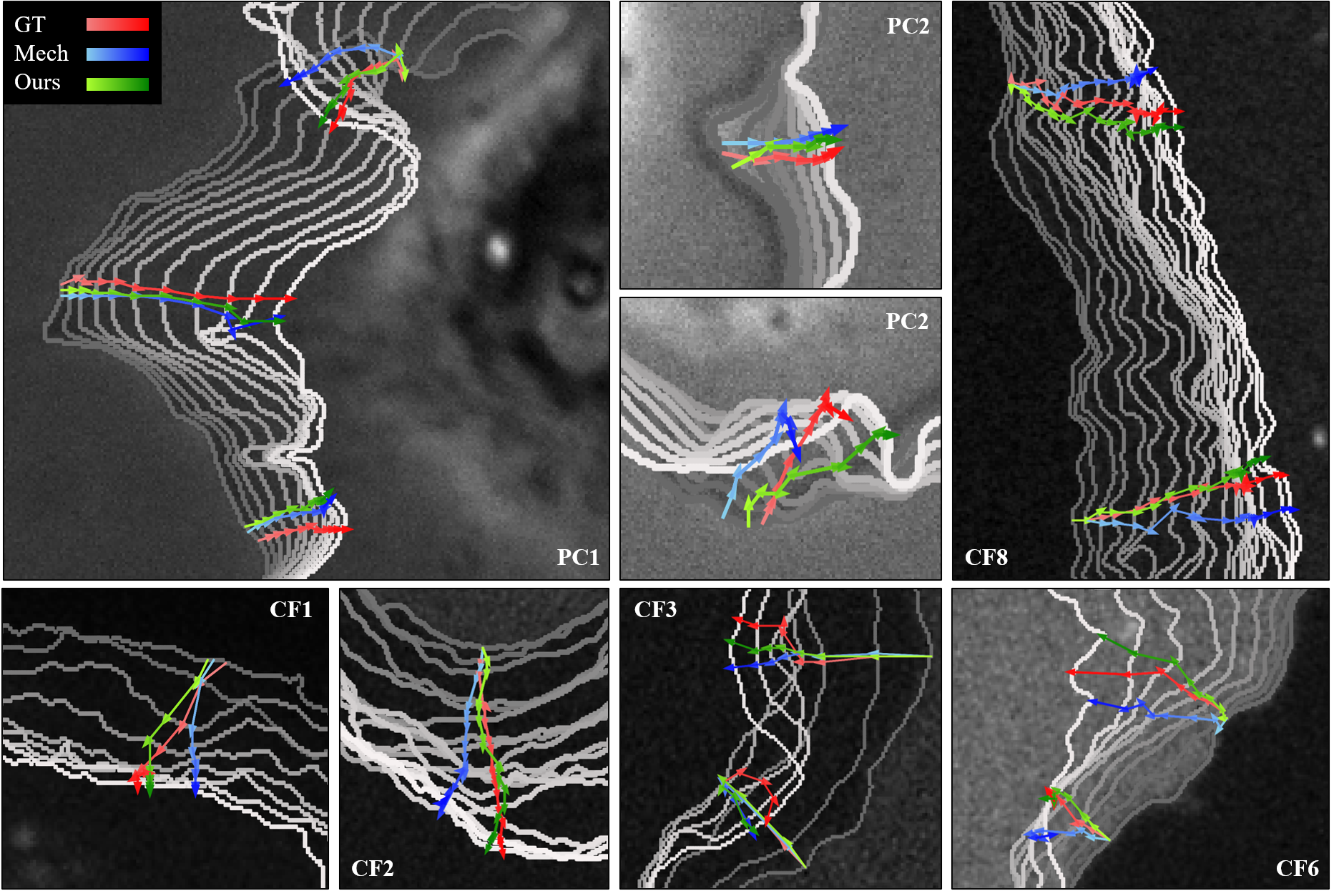}
   \caption{\textbf{Trajectories of tracked points by our contour tracker (green), the mechanical model (blue), and ground truth labels (red).} To indicate the time from the initial frame to the last frame, the color of the trajectory gradually changes from light to dark, and the color of the contour changes from gray to white. PC prefix refers to phase contrast live cell dataset \cite{MARS_Net} and CF prefix refers to confocal fluorescence live cell dataset \cite{Wang2018}. The number refers to the video number. Trajectories in PC1, PC2, and CF1 start from the mid-frame, and the rest of the trajectories start from the first frame at the same initial point.}
   \label{fig:prediction_result-fig}
    \vspace{-0.4cm}
\end{figure*}


\subsection{Ablation Study}
The spatial accuracy (SA) and contour accuracy (CA) are measured with multiple thresholds since some models can perform better in lower thresholds while others perform better in higher thresholds. 

\noindent \textbf{Loss Functions.}\quad
We compare one supervised learning loss and combinations of four unsupervised learning losses in \cref{tab:loss ablation table}. 
For supervised learning on dense tracking points, contour point-to-point correspondences predicted by the mechanical model \cite{Machacek2006} are used as ground truth correspondences. 
Then, our contour tracker is trained to minimize the L2 distance between predicted points' and ground truth points' locations, similar to PoST \cite{nam2021polygonal}.
The same architecture shown in \cref{fig:architecture-fig} is used for ablation.
Supervised learning yields lower accuracy than training with mechanical-normal loss or cycle consistency loss alone.
The low accuracy can be due to inaccurate pseudo-labels or overfitting on training labels which does not generalize to new contour points in validation videos.

Unsupervised learning by the mechanical-normal loss or cycle consistency loss alone have significantly higher performance than other losses, such as photometric loss. 
However, adding the mechanical-linear loss to the mechanical-normal loss degrades the performance. 
Training with the mechanical-linear loss alone or other combinations of losses not shown in the table also yields low accuracy.
Adding the mechanical-normal loss and cycle consistency loss yields the highest spatial and contour accuracy.

\noindent \textbf{Architecture.}\quad
We replaced or removed a component of our architecture to see how each component contributes to the overall performance, as shown in \cref{tab:architecture ablation table}.
No cross attention (No Cross) fuses point features sampled from the first and the second images by adding them.
Single cross attention (Single Cross) only uses one cross attention to fuse point features. 
The contour tracker with one cross attention outperforms the contour tracker without any cross attention in a low threshold setting but not in higher threshold settings.
Single cross attention is not effective, possibly because the movement of the live cell played backward is physically different than the natural live cell movement. 
So contour features need to be handled differently depending on forward or backward directions.
Our contour tracker using forward and backward cross attentions yields much higher accuracy than using one or zero cross attention. 

From DeepSnake \cite{peng2020deep} and PoST \cite{nam2021polygonal}, circular convolution is known to be effective for point regression given a sequence of contour points. 
When FCNN is replaced with circular convolution (Circ Conv), the model achieves much lower accuracy at low thresholds.
Since our contour tracker handles a cellular contour with disconnected endpoints, we also test 1D convolution  (1D Conv).
Using circular convolution or 1D convolution decreases the accuracy at low thresholds.
We chose the model with the highest accuracy at low thresholds because the model's accuracy at low thresholds reveals its pixel-level accuracy, and high pixel-level accuracy is known to yield less noisy and stronger morphodynamic patterns \cite{MARS_Net}. 

\subsection{Comparison with Other Methods}
We compare our contour tracker against mechanical model \cite{Machacek2006} and other deep learning-based methods: UFlow \cite{Jonschkowski2020matters} and PoST \cite{nam2021polygonal}. Since pre-trained PoST without any modifications yields very low spatial and contour accuracy, we modified it to utilize features along the current $C_{t}$ and the next contours $C_{t+1}$. Also, we trained PoST on the live cell dataset with our cycle consistency loss. For inference, both UFlow and PoST offset points $O_{t \rightarrow t+1}(p_t)$ move to the closest contour points in $C_{t+1}$. This is the same inference heuristic used for our contour tracker.

\noindent \textbf{Phase Contrast Dataset.}\quad Our contour tracker outperforms all the other methods in all threshold settings in \cref{tab:MARS-Net table}.
UFlow\cite{Jonschkowski2020matters} has trouble tracking contour points with the lowest accuracy. Unlike other methods, UFlow does not use contour features and does not focus on tracking the contour only. 
PoST \cite{nam2021polygonal} performs better than UFlow but still worse than the mechanical model or our contour tracker.
Lastly, the mechanical model performs better than other deep learning models except ours. 

\noindent \textbf{Confocal Fluorescence Dataset.}\quad
The overall performance in \cref{tab:HACKS table} is lower than \cref{tab:MARS-Net table} because the confocal fluorescence dataset \cite{Wang2018} is taken in higher resolution with fine details and contains some segmentation error from thresholding.
Despite the difficulty, our contour tracker still outperforms all the other methods in all threshold settings as shown in \cref{tab:HACKS table}.
Consistent with the quantitative results, we qualitatively show in \cref{fig:prediction_result-fig} that our contour tracker can track contour points with closer proximity to the ground truth labels compared to the mechanical model in both phase contrast and confocal fluorescence datasets. 
Please refer to the supplementary section for the visualization of a long sequence of videos.

\begin{figure}[t]
  \centering
   \includegraphics[width=1.0\linewidth]{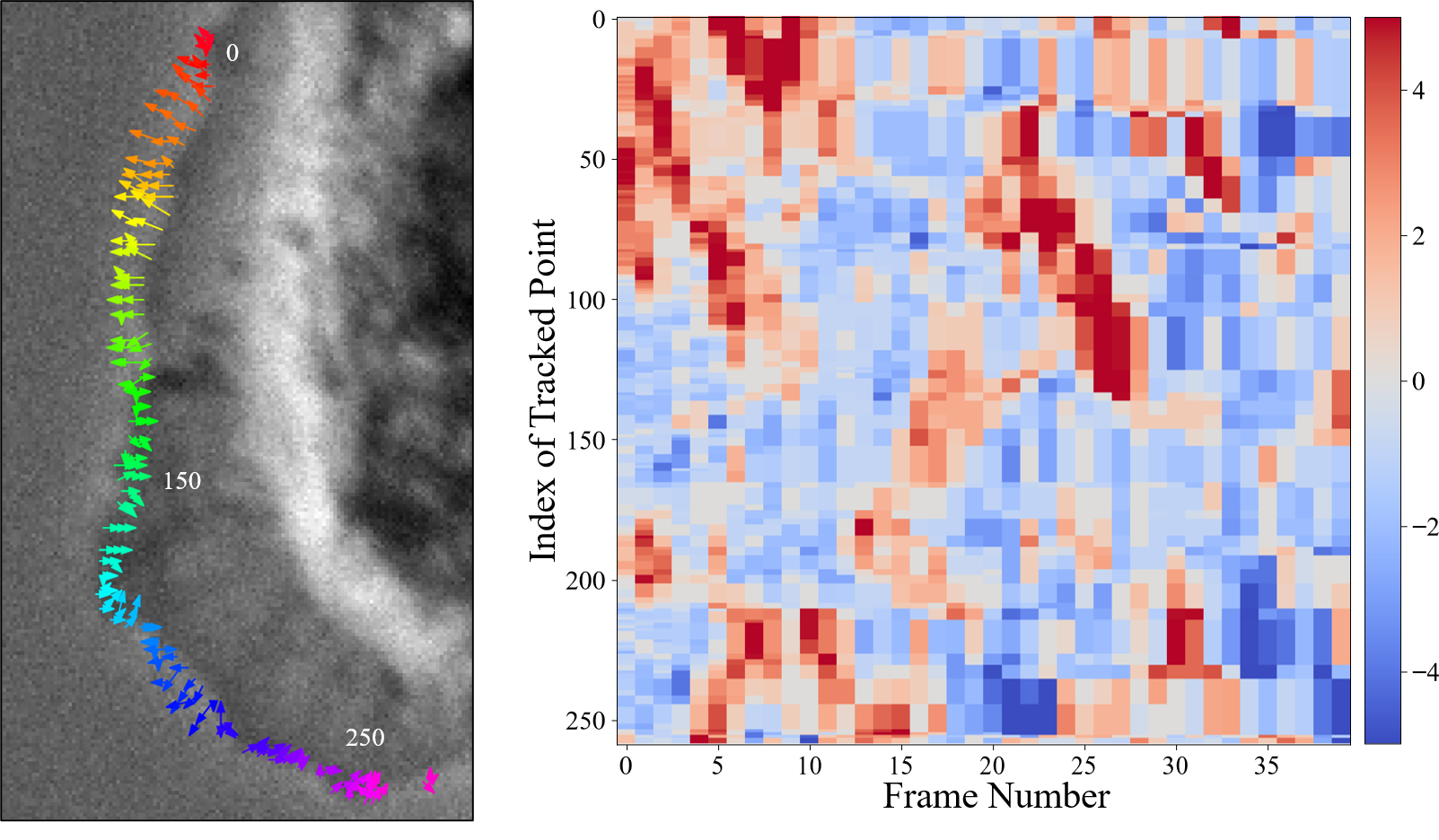}

   \caption{\textbf{Quantification of morphodynamics from our contour tracking results.} 
   Trajectories of tracked contour points for 3 frames on a phase contrast live cell \cite{MARS_Net} are shown on the left, and the quantification of those tracked points as velocities for 40 frames are shown on the right.
   }
   \label{fig:morphodyanmics-fig}
   \vspace{-0.4cm}
\end{figure}

\begin{figure}[t]
  \centering
   \includegraphics[width=1.0\linewidth]{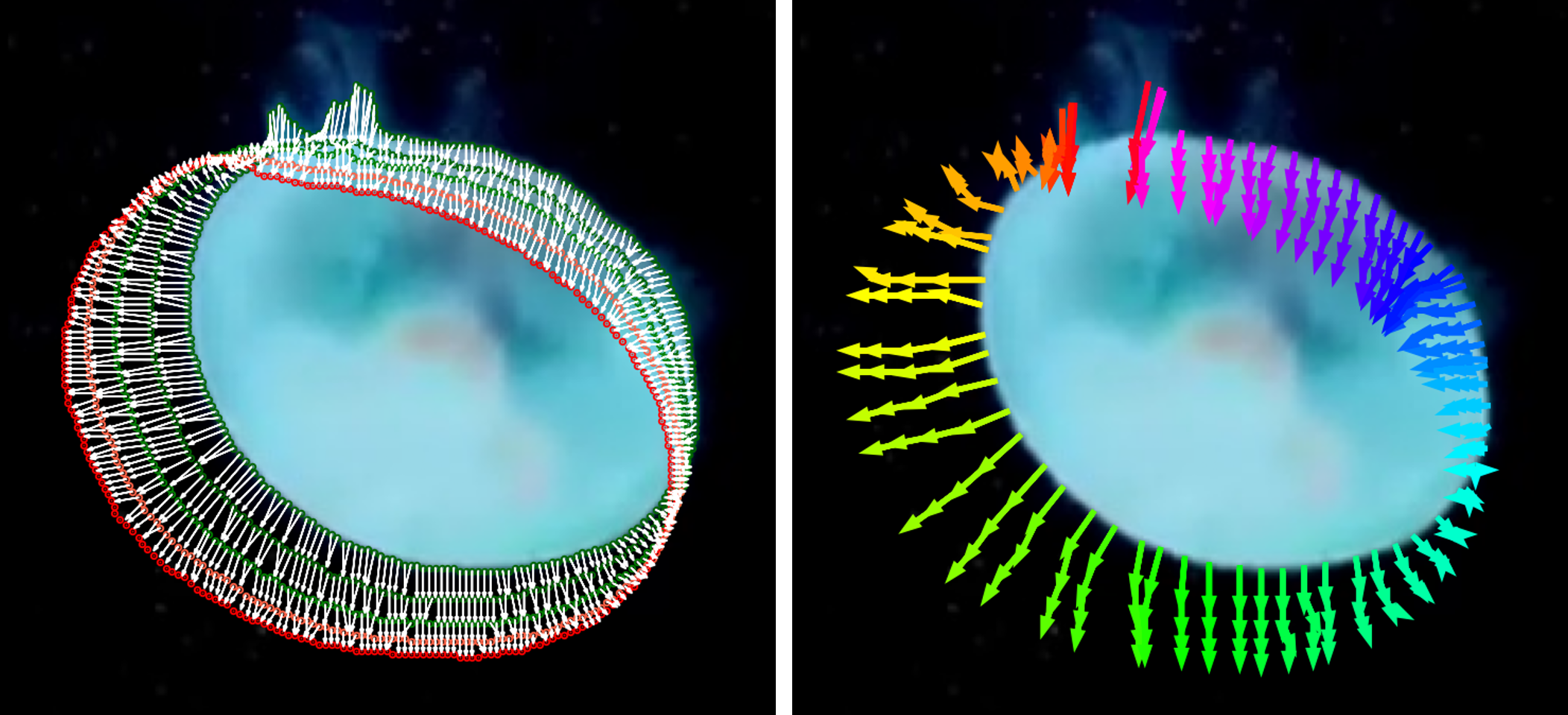}

   \caption{\textbf{Visualization of contour tracking results on a jellyfish \cite{skorokhodov2022stylegan}.} Dense correspondences between adjacent contours with white arrows are shown on the left. The color of the contour changes from green to red as the frame number increases. The trajectories of a few tracked points are shown on the right.
   }
   \label{fig:jellyfish-fig}
   \vspace{-0.3cm}
\end{figure}


\subsection{Quantification of Morphodynamics}
We quantify cellular morphodynamics of one of the phase contrast live cell videos \cite{MARS_Net} tracked by our contour tracker as a heatmap in \cref{fig:morphodyanmics-fig}.
We chose two far-apart contour points such that the velocities of all contour points between those two points are measured.
Only the velocity along the normal vector of contour points is considered.
The red regions indicate outward motion (protrusion) from the cell body and the blue regions indicate inward motion to the cell nucleus of the cellular contour.

\subsection{Contour Tracking of a Jellyfish}
Our live cell videos contain live cells anchored to one of the sides of the image. 
In this section, we show that our contour tracker can also work on a different viscoelastic organism floating in space.
We observed that jellyfish has a similar viscoelastic property to live cell, so we tested our contour tracker on Rainbow Jellyfish Benchmark from StyleGan-V \cite{skorokhodov2022stylegan}.
We cropped the center of a video to get $512^2$ patches containing a jellyfish and segmented it by thresholding. 
Our contour tracker can track the contour of a jellyfish with dense point-to-point correspondences as shown in \cref{fig:jellyfish-fig}. 
Please refer to the supplementary section for more details.

\section{Conclusion}
We present a novel deep learning-based contour tracking with correspondence for live cells and train it without any ground truth tracking points. We systematically tested various unsupervised learning strategies on top of the proposed architecture with cross attention and found that a combination of mechanical and cycle consistency losses is the best. 
Our contour tracker outperforms the classical mechanical model and other deep learning-based methods on phase contrast and confocal fluorescence live cell datasets. 
In the field of computer vision, we hope this work sheds light on a new type of object tracking (e.g. viscoelastic materials), which prior arts cannot adequately capture.
\\

\noindent \textbf{Acknowledgements.}\quad
J. Jang and T-K Kim are in part sponsored by NST grant (CRC 21011, MSIT), KOCCA grant (R2022020028, MCST), and IITP grant funded by the Korea government(MSIT)(No.2019-0-00075, Artificial Intelligence Graduate School Program(KAIST)). K. Lee is supported by NIH, United States (Grant Number: R35GM133725).
\newpage
{\small
\bibliographystyle{ieee_fullname}
\bibliography{egbib}
}

\newpage

\renewcommand{\thesection}{S.\arabic{section}}
\section*{S. Supplementary Material}

\setcounter{section}{0}
\section{Addtional Methods}

\noindent \textbf{Mechanical-Linear Loss.}\quad Similar to linear spring force from the mechanical model \cite{Machacek2006}, the mechanical-linear loss measures the average distance $A=\sum_{i=0}^{N-1} \| p^{i+1}_{t+1} - p^{i}_{t+1} \|_2$ between adjacent predicted contour points and forces every distance between adjacent points to be the same as the average distance. Here, each predicted contour point is $p^{i}_{t+1} = \phi(O_{t \rightarrow t+1} (p^i_t))$. The mechanical-linear is different from Snake \cite{kass1988snakes} algorithm's tension term, which simply minimizes the distance between adjacent points.
\begin{equation}
  L_{\text{mech-linear}} = \sum_{i=0}^{N_t-2} \| \| p^{i+1}_{t+1} - p^{i}_{t+1} \|_2 - A \|_1
  \label{eq:linear spring loss}
\end{equation}

\noindent \textbf{Photometric Loss.}\quad
Inspired by UFlow \cite{Jonschkowski2020matters}, we implemented photometric loss that minimizes the forward-then-backward tracked point's pixel intensity to match its original point's pixel intensity. Similar to the cycle consistency loss, forward and backward consistency losses are combined. But photometric loss measures the difference in pixel intensity, not the distance.
\begin{equation}
\begin{aligned}
  L_{photo}&=\sum_{i=0}^{N_t-1} \| \gamma(p^i_t) - \gamma(O_{t+1\rightarrow t}(p^{i}_{t+1})) \|_1 \\
   &+ \sum_{i=0}^{N_t-1} \| \gamma(p^i_{t+1}) - \gamma(O_{t\rightarrow t+1}(p^{i}_{t})) \|_1
  \label{eq:photometric loss}
\end{aligned}
\end{equation}

Retrieving a point's pixel intensity is denoted as $\gamma$. 

\noindent \textbf{Segmentation}\quad
To provide an ordered sequence of contour points $C_t$ as input to our contour tracker, the cell body is segmented first.
Confocal fluorescence live cell videos \cite{Wang2018} have a distinct boundary between the cell body and dark background, so conventional image thresholding is enough to obtain their segmentation masks. However, the live cell videos taken with phase contrast microscope \cite{MARS_Net} have challenging visual features such as halo and shade-off artifacts. As a result, a specialized deep segmentation model \cite {MARS_Net} for phase contrast live cell videos is adopted. Both datasets used in our paper are not manually segmented, so their segmentation masks contain some noise.

\noindent \textbf{Labeling Tool}\quad
We developed the labeling tool in Python to expedite the labeling process, as shown in \cref{fig:labeling_GUI}. It is available as an executable application in Windows 10/11. The labeling tool loads images with contour tracking points marked on the contour of the image. 
A user can click on the main view to create a tracking point, indicated by the blue circle. 
Scrolling up or down zooms in or out at the location of the user's cursor. The zoomed-in window is shown on the left pane of the GUI.
Prev or next button on the bottom changes the frame such that the user can view how points move in consecutive frames and label tracking points. Created tracking points can be dragged to change their locations.
Press the save button to save tracking points labeled at the current frame, or press the clear button to remove all labeled tracking points.

\noindent \textbf{Training Details}\quad
In this section, we provide training details about our contour tracker and two compared methods, UFlow \cite{Jonschkowski2020matters}, and PoST \cite{nam2021polygonal}. 
Our contour tracker, UFlow and PoST are all trained from scratch except for the ImageNet \cite{deng2009imagenet} pre-trained VGG16 encoder and ResNet50 encoder in our contour tracker and PoST, respectively.
The architecture of UFlow \cite{Jonschkowski2020matters} is kept the same, but we modified PoST \cite{nam2021polygonal} to improve its performance since their original architecture did not perform well in our dataset. 
Our contour tracker, UFlow \cite{Jonschkowski2020matters}, and PoST \cite{nam2021polygonal} are trained by unsupervised learning with cycle consistency loss. But our contour tracker is also trained with mechanical-normal loss. 
UFlow \cite{Jonschkowski2020matters} only takes images as input, while PoST \cite{nam2021polygonal} and our contour tracker take both images and contour points as input for training and inference.

\counterwithin{figure}{section}
\setcounter{figure}{0}

\begin{figure}[t]
  \centering
  \includegraphics[width=1.0\linewidth]{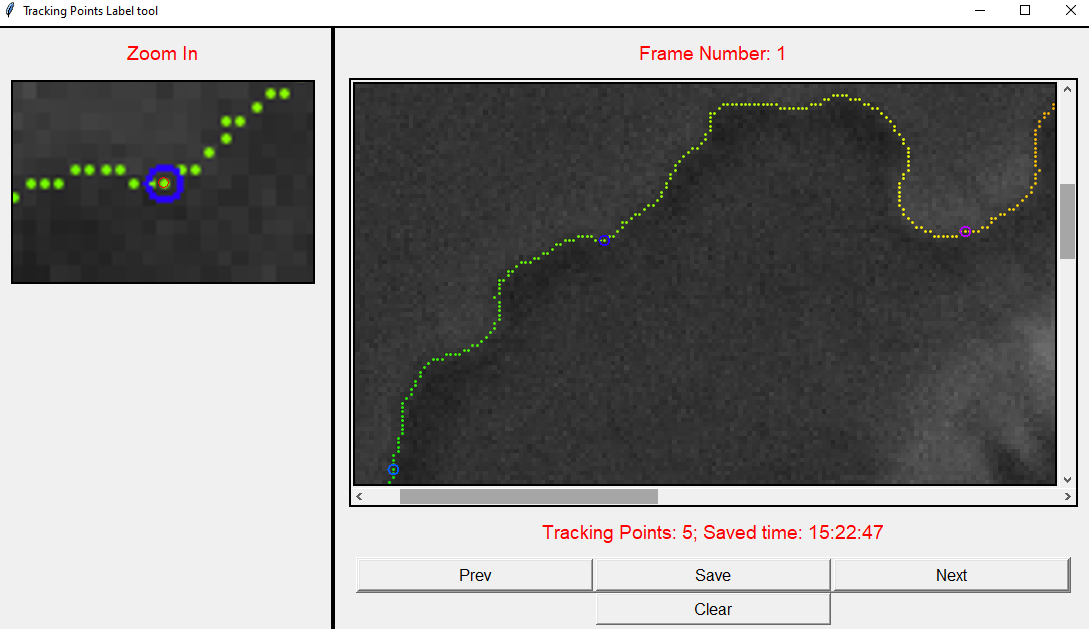}
   \caption{GUI for labeling tracking points.}
   \label{fig:labeling_GUI}
\end{figure}

\section{Addtional Results}
\label{sup_livecell}
We qualitatively validate dense point-to-point correspondences in the long sequences of phase contrast (PC) \cite{MARS_Net} and confocal fluorescence (CF) live cell videos \cite{Wang2018} and a jellyfish video \cite{skorokhodov2022stylegan} on our webpage at {\small \url{https://junbongjang.github.io/projects/contour-tracking/}}. They are best viewed in 1080p, HD resolution.
The dense correspondences are shown with arrows pointing from the first frame's contour $C_t$ to the second frame's contour $C_{t+1}$. 
The arrows are colored by the sequence of contour points' order from red to purple.
The length of the arrow is the magnitude of the contour point's movement.
The direction and magnitude of the arrows align well with our expectations, except for some abrupt changes in contour points due to the segmentation error.

For the jellyfish video, we picked 200 frames long sequence containing a single jellyfish from the Rainbow Jellyfish Benchmark \cite{skorokhodov2022stylegan}.
Since our contour tracker is trained by unsupervised learning in less than a day, we train on a jellyfish video and predict dense point-to-point correspondences on the same video, which removes the need for making a training dataset.

\begin{figure}[t]
  \centering
  \includegraphics[width=1.0\linewidth]{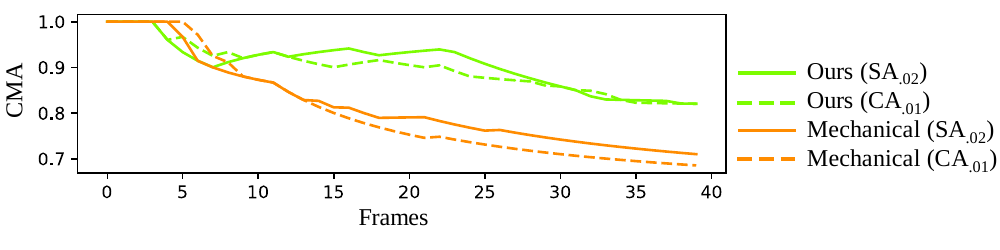}
   \caption{Cumulative mean accuracy (CMA) from the first to the last frame on one of the confocal fluorescence videos \cite{Wang2018}.}
   \label{fig:accuracyplot}
\end{figure}

\noindent \textbf{Short \& Long Term Tracking.}\quad
In our main paper, we only evaluated long-term tracking of the points from the first to the last frame, so the point tracked from the previous frame was used to track in the next frames. Additionally, we evaluate both short-term and long-term tracking in \cref{fig:accuracyplot}, which shows that the tracking error accumulates as the number of frames to track increases. At first, the Mechanical model \cite{Machacek2006}, and our contour tracker has similar tracking accuracy, but the Mechanical model's \cite{Machacek2006} tracking accuracy decreases much faster than our contour tracker's.
Cumulative Mean accuracy is obtained by averaging all SA or CA up to the frame number $t$, which is the x-axis in \cref{fig:accuracyplot}.

\noindent \textbf{Error Study.}\quad
We performed an error study in the phase contrast videos \cite{MARS_Net}. 
The tracking accuracy is linearly proportional to the magnitude of the cellular movement.
The spatial accuracy (SA$_{.02}$) decreases by about 0.02 as the absolute velocity increases by 1.
The velocity is in unit pixels/frame and is perpendicular to the cellular contour.
During cellular expansion and contraction, SA$_{.02}$ is 0.731 and 0.710, respectively, so the direction of the cellular motion does not significantly affect the accuracy.

\noindent \textbf{Forward and Backward Tracking.}\quad
Only the forward cross attention is used to regress offsets for inference because forward tracking achieves significantly higher accuracy than backward tracking by 0.16/0.43 at SA$_{.02}$/CA$_{.01}$ in the phase contrast videos \cite{MARS_Net}. 
Ideally, both forward and backward tracking's performance should be similar, but our contour tracker is trained with mechanical-normal loss that updates the weights of the forward cross attention layer only during backpropagation.
For computational efficiency during inference, the backward cross attention layer can be removed from our contour tracker such that backward offsets $O_{t+1 \rightarrow t}$ are not predicted.

\section{Limitations}
The segmentation error affects the performance of both the mechanical model and our contour tracker, which uses features from the contour and maps correspondence between contour points. 
For future work, jointly training our contour tracker with the segmentation model end-to-end will refine contours and then find contour point correspondences.
Also, segmentation accuracy can be further improved by transfer learning with diverse microscopy datasets \cite{MARS_Net} or a human-in-the-loop approach without preparing large-scale datasets \cite{pachitariu2022cellpose}.

Our contour tracker can naturally handle many-to-one correspondences (merging) but not one-to-many correspondences (splitting). For future work, implementing one-to-many correspondences should handle expanding contours even better. 
One potential way is to regress backward offset from the second contour to the first contour and find many-to-one correspondences, which become one-to-many correspondences when reversed.

\end{document}